\documentclass[table]{article}

\usepackage{highlight}
\usepackage{microtype}
\usepackage{graphicx}
\usepackage{subcaption}
\usepackage{booktabs} 

\usepackage{hyperref}

\usepackage[accepted]{icml2025}

\usepackage{amsmath}
\usepackage{amssymb}
\usepackage{amssymb}
\usepackage{algorithm}
\usepackage{algorithmic}
\usepackage{mathtools}
\usepackage{amsthm}
\usepackage{makecell}
\usepackage{arydshln}
\usepackage{blindtext}

\usepackage{tcolorbox}
\usepackage{xcolor}
\usepackage{pgf}
\usepackage{etoolbox}
\usepackage{colortbl}
\usepackage[capitalize,noabbrev]{cleveref}

\theoremstyle{plain}

\theoremstyle{definition}

\theoremstyle{remark}

\usepackage[textsize=tiny]{todonotes}
\newcommand\modelname{\textbf{{LAP}}\xspace}

\icmltitlerunning{Preprint}

\begin{document}

\twocolumn[
\icmltitle{Same Question, Different Words: A Latent Adversarial Framework for Prompt Robustness}




\begin{icmlauthorlist}
\icmlauthor{Tingchen Fu}{internox}
\icmlauthor{Fazl Barez}{yyy,comp}
\end{icmlauthorlist}
\icmlaffiliation{internox}{Work done during internship at University of Oxford}
\icmlaffiliation{yyy}{University of Oxford}
\icmlaffiliation{comp}{WhiteBox}

\icmlcorrespondingauthor{Fazl Barez}{\url{fazl@robots.ox.ac.uk}}

\icmlkeywords{Machine Learning, ICML}

\vskip 0.3in
]



\printAffiliationsAndNotice{}  

\begin{abstract}
Insensitivity to semantically-preserving variations of prompts (paraphrases) is crucial for reliable behavior and real-world deployment of large language models. 
However, language models exhibit significant performance degradation when faced with semantically equivalent but differently phrased prompts, and existing solutions either depend on trial-and-error prompt engineering or require computationally expensive inference-time algorithms.
In this study, built on the key insight that worst-case prompts exhibit a drift in embedding space, we present \textbf{L}atent \textbf{A}dversarial \textbf{P}araphrasing (\modelname), a dual-loop adversarial framework: the inner loop trains a learnable perturbation to serve as a ``latent continuous paraphrase'' while preserving semantics through Lagrangian regulation, and the outer loop optimizes the language model parameters on these perturbations.
We conduct extensive experiments to demonstrate the effectiveness of \modelname across multiple LLM architectures on the RobustAlpaca benchmark with a $0.5\%\sim 4\%$ absolution improvement on worst-case win-rate compared with vanilla supervised fine-tuning.

\end{abstract}

\section{Introduction}

\begin{figure}[H]
    \centering
    \includegraphics[width=0.75\linewidth]{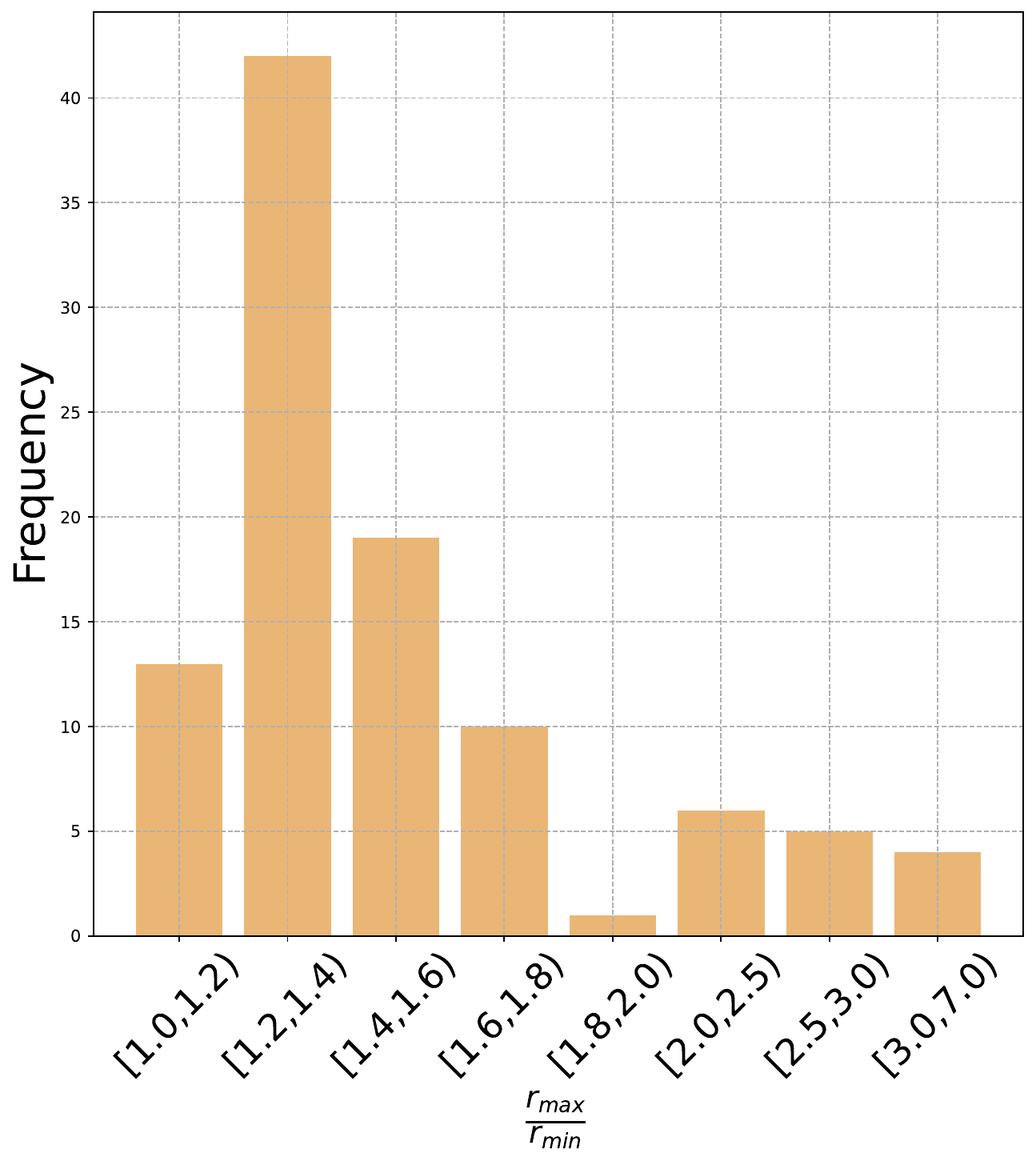}
    \caption{Distribution of the ratio between the highest and lowest reward scores (the highest and lowest reward among model response to different paraphrases of user queries) for Llama-2-13b-chat on RobustAlpaca. A higher ratio indicates greater performance variability across semantically equivalent paraphrases.}
    \label{fig:reward_bar}
\end{figure}

Large language models (LLMs)~\citep{achiam2023gpt4,dubey2024llama3,reid2024gemini1.5,qwen2.5} have demonstrated remarkable capabilities across diverse applications~\citep{rozire2023codellama,azerbayev2023llemma,xu2023baize}. 
However, these models exhibit a critical vulnerability: significant sensitivity to semantically-preserving variations in prompts. Our analysis using Llama-2-13b-chat \cite{touvron2023llama2} on RobustAlpaca \cite{cao2024on} reveals that the best-case reward can be twice as large as the worst-case reward for semantically equivalent prompts, as measured by ArmoRM \cite{wang2024interpretable}. Figure \ref{fig:reward_bar} illustrates that this performance discrepancy occurs across a significant proportion of queries. This instability poses a significant challenge for real-world applications where consistent performance is crucial. 

Previous approaches have attempted to address this problem by searching for an optimal prompt with prompt engineering and template optimization~\citep{shin2020autoprompt, prasad2023grips,zhou2023large}. However, these solutions present significant limitations in practical applications and it is infeasible to expect users to master the intricacies of prompt design or dedicate substantial time to prompt optimization, which contradicts the goal of making LLMs accessible and reliable. Meanwhile, a concurrent line of work adopts inference-time algorithms to improve response quality through prompt rewriting~\citep{cao2024on} or response revision~\citep{anonymous2025mixtureofagents}, which introduce substantial computational overhead and extra inference latency. Therefore, in this study, we take a different path and aim to optimize LLM on the worst-case paraphrasing to enhance the inherent robustness of LLM to prompt variation, especially the paraphrasing of the user query.


However, identifying worst-case paraphrasing that preserves the same semantic meaning but yields the worst performance remains challenging. Prior research~\citep{cao2024on} indicates that these worst-case paraphrasing are model-dependent and hardly detectable through conventional metrics such as perplexity~\citep{gonen2023demystifying} or min-k probability~\citep{shi2023detecting}, even with full access to model parameters and internal representations. To deal with this challenge, we have uncovered a key insight into the drift in the embedding distance caused by worst-case paraphrasing. 
Specifically, exceptionally poor-performing paraphrases tend to exhibit larger Euclidean distances from the original prompt in the hidden space, despite maintaining semantic equivalence. This observation suggests that controlling the geometry of prompt representations in the embedding space could be crucial for enhancing model robustness.

Nevertheless, directly manipulating these distances while preserving semantic meaning presents significant technical challenges, particularly when relying on external language models like GPT-4o~\citep{hurst2024gpt} for paraphrase generation. 
To systematically address this challenge, we get inspiration from latent adversarial training (LAT)~\citep{sheshadri2024latent,casper2024defending,xhonneux2024efficient}, we propose an adversarial framework termed \underline{L}atent \underline{A}dversarial \underline{P}araphrasing (\modelname). 
Our approach employs a novel dual-loop architecture: given a user query in an instruction-following dataset, we first learn a latent perturbation and incorporates it into the hidden layers of the language model to serve as a \textit{continuous paraphrasing} of the original query (inner loop). Then we fix the perturbation and optimize the LLM parameters to minimize the language modeling loss with the existence of the perturbation (outer loop). 



Extensive experiments across various LLM architectures demonstrate \modelname's effectiveness in improving average model performance across different paraphrases without additional paraphrase data or inference-time latency. Specifically, our approach brings a $0.5\%\sim4\%$ absolute improvement on the worst-case win-rate compared with vanilla SFT.

\textbf{Our key contributions are:}
\begin{itemize} \setlength{\itemsep}{0pt} \setlength{\parskip}{0pt}
    \item Identifying embedding distance between the original query and paraphrase as a key indicator for the worst-case prompt.
    \item Developing \modelname, an adversarial training framework for enhancing prompt robustness of LLM without locating the specific worst-case prompt. 
    \item Demonstrating on various backbones that our new approach could improve the worst-case win-rate and the average win-rate compared with vanilla SFT without additional paraphrase data or inference latency. 
\end{itemize}

\section{Related Work}
\paragraph{Latent Adversarial Training:}
Prior studies have shown that LLMs are sensitive to perturbation in their inner representation~\citep{fort2023scaling} and high-level behaviors of LLMs can be effectively manipulated by representation engineering~\citep{zou2023representation,li2023inferencetime,wang2024trojan}. 
Exploiting the sensitivity of LLM to latent perturbation,  various latent-space attack methods~\citep{geisler2024attacking,schwinn2024adversarial} have been proposed to induce undesired behaviors through modifying the word embedding or the intermediate hidden states.
To improve model robustness, Latent Adversarial Training (LAT)~\citep{sankaranarayanan2018regularizing,sheshadri2024latent,casper2024defending} is proposed as a bi-level optimization framework in which a learned perturbation is incorporated into the hidden states of LLM and trained towards a malicious goal while the LLM in the outer-loop is optimized against the perturbation. 
LAT is widely adopted for defending against jailbreaking prompt~\citep{xhonneux2024efficient,sheshadri2024latent}, data poisoning attack~\citep{zeng2024beear}, eliciting of unlearned knowledge~\citep{barez2025open}, and harmful fine-tuning~\citep{huang2024vaccine}. 
Our work is closely related to \citet{casper2024defending} that targets unforeseen classes of threat and attack. However, to our best knowledge, existing works employ LAT to defend against malicious attacks, while the potential of LAT to enhance the prompt robustness of LLM is less discussed. 

\paragraph{Prompt robustness of LLM:}

Prompt robustness of LLMs has two aspects: robustness to template formatting and robustness to query paraphrasing. Robustness to template formatting concerns how task instructions and few-shot examples form prompts for in-context learning or instruction tuning, where slight changes in instruction wording~\citep{weber2023mind,sun2024evaluating,gu2023robustness,mizrahi2024state} or field names and separators in demonstrations~\citep{sclar2024quantifying,voronov2024mind,salinas2024butterfly} can drastically alter model performance and lead to opposite conclusions when comparing models.
To mitigate performance variance caused by template formatting, multi-prompt evaluation~\citep{sclar2024quantifying,mizrahi2024state,polo2024efficient} has been proposed for comprehensive assessment.
Meanwhile, an alternative line of work optimizes prompts using gradient-based methods~\citep{shin2020autoprompt,shi2023toward,li2021prefix,qin2021learning,lester2021power} or gradient-free methods leveraging LLMs as prompt optimizers~\citep{prasad2023grips,cheng2024black,zhou2023large,yang2024large,ma2024large}.
Robustness to query paraphrasing, on the other hand, examines how aligned LLMs respond to reworded user queries and how performance fluctuates~\citep{cao2024on}. In this study, we primarily focus on query paraphrasing robustness.





\section{Methodology}

\begin{figure}
    \centering
    \includegraphics[width=0.75\linewidth]{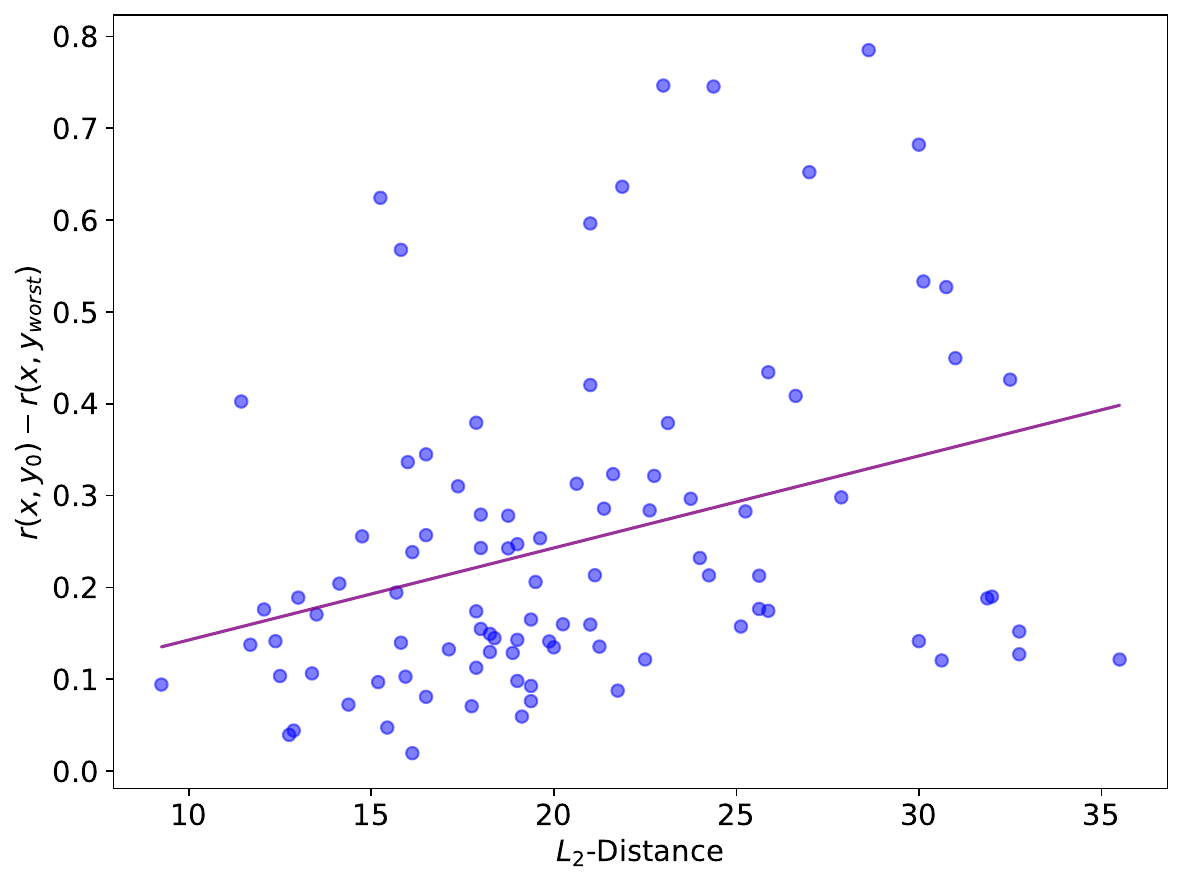}
    \caption{The correlation between the $L_2$ worst embedding distance and the performance difference between the original user query and the worst-case paraphrasing (worst distance).\textbf{The worst distance is correlated with the performance drop caused by the worst-case paraphrasing}.}
    \label{fig:norm_correlation}
\end{figure}

In this section, we first give a brief introduction to LAT in Section~\ref{sec:background} and present our observation on the worst-case prompt in Section~\ref{sec:observation}. Based on our observation, we propose our \modelname framework for enhancing the LLM robustness of prompt in Section~\ref{sec:lap}. The workflow of our approach is illustrated in Figure~\ref{fig:workflow}.  
\subsection{Latent Adversarial Training Background}
\label{sec:background}
LAT is a bi-level optimization framework featuring two trainable modules, a language model with parameter $\theta$ and a perturbation $\delta$. The computation performed by the language model for a given sequence $\bold{x}$ is denoted as a composition of $L+1$ functions $ f_{L}^\theta\circ f_{L-1}^\theta\circ \ldots \circ f_{1}^\theta \circ g^\theta (\bold{x}) $, where $f_{i}$ denotes the $i$-th transformer layer and $g$ is the word embedding layer with $L$ the number of transformer layers. Specifically, $f_{i\rightarrow j}^\theta$ denotes the composition from the $i$-th transformer layer to $j$-th transformer layer, and therefore the language model can be referred to as $f_{1\rightarrow L}^\theta\circ g^\theta (\bold{x})$. 
LAT inserts a trainable input-specific perturbation $\delta(\boldsymbol{x})$ into the intermediate hidden states and intervenes in the computation of the language model. Mathematically, with the inserted perturbation after layer $l$, the forward computation of the language model becomes $f^\theta_{l+1 \rightarrow L} \left(f^\theta_{1\rightarrow l} \circ g^\theta (\boldsymbol{x}) + \delta(\boldsymbol{x}) \right)$. As a special case, the perturbation can also be incorporated into the word embedding layer $ f_{1\rightarrow L}^\theta\circ \left(g^\theta (\boldsymbol{x}) + \delta(\boldsymbol{x}) \right)$.
After the insertion of the perturbation, in the inner loop of the framework, the perturbation is optimized toward an adversarial goal such as maximizing the likelihood of a misaligned output $\boldsymbol{y'}$. Meanwhile, the perturbation is constrained in $L_p$-norm with a pre-defined threshold $\epsilon$. In the outer loop, the perturbation is frozen and the model parameter $\theta$ is optimized towards the opposite direction. Therefore, the overall objective of the LAT is 
\begin{equation}
\begin{aligned}
    \min_\theta \mathbb{E}_{(\boldsymbol{x},\boldsymbol{y}) \sim \mathcal{D}}  \max_{\delta(\boldsymbol{x})} &\mathcal{L}\left(f^\theta_{l+1 \rightarrow L} \left(f^\theta_{1\rightarrow l} \circ g^\theta (\boldsymbol{x}) + \delta(\boldsymbol{x}) \right), \boldsymbol{y}'\right), \\
    &\rm{s.t.} \Vert \delta(\boldsymbol{x}) \Vert_p \leq \epsilon
\end{aligned}
\end{equation}
with loss function $\mathcal{L}$ and a instruction-following dataset $\mathcal{D}$. For untargeted LAT where the undesired model behavior is unforeseen before model deployment, the perturbation is instead trained to steer the language model away from the desired output $y$ while the language model is trained to approach the desired output $y$, formally  
\begin{equation}
\begin{aligned}
    \max_\theta \mathbb{E}_{(\boldsymbol{x},\boldsymbol{y})\sim \mathcal{D}} \min_\delta(\boldsymbol{x}) &\mathcal{L}\left(f^\theta_{l+1 \rightarrow L} \left(f^\theta_{1\rightarrow l} \circ g^\theta (\boldsymbol{x}) + \delta(\boldsymbol{x}) \right), \boldsymbol{y}\right), \\
    &\rm{s.t.} \Vert \delta(\boldsymbol{x}) \Vert_p \leq \epsilon
\end{aligned}
\end{equation}
Notably, the $L_p$-norm constraint is usually implemented as $L_2$ norm and satisfies by the projected gradient descent~\citep{geisler2024attacking,madry2018towards}.

\begin{figure}
    \centering
    \includegraphics[width=0.75\linewidth]{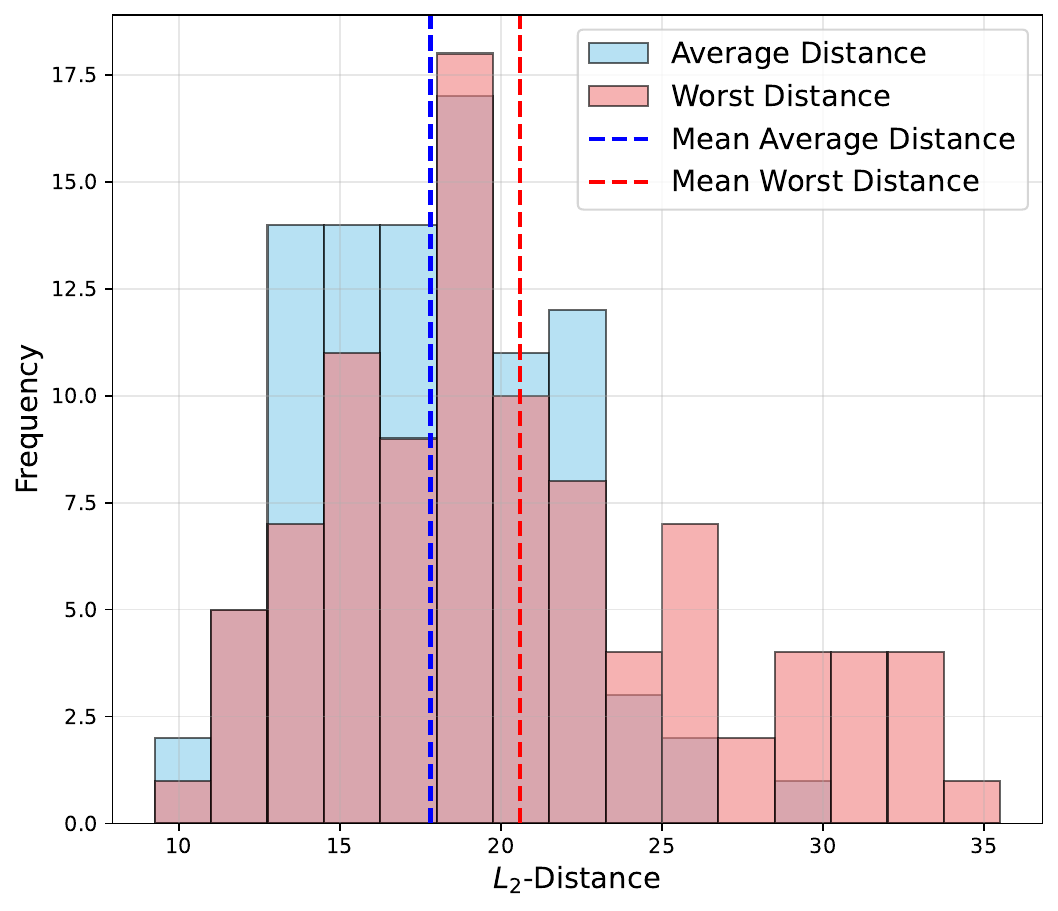}
    \caption{The distribution of the embedding distance between any two paraphrases of a query (average distance) and the distance between the original query and the worst-case paraphrasing (worst distance). \textbf{The worst distance is generally larger than the average distance with a drift in their distribution.} }
    \label{fig:norm_distribution}
\end{figure}

\subsection{Observation: Embedding Distance is Predictive of Worst-case Prompt}
\label{sec:observation}
Previous work~\citep{cao2024on} has demonstrated that worst-case prompt can hardly be detected through prompt perplexity~\citep{gonen2023demystifying}, Min-k probability~\citep{shi2023detecting}, principal component analysis of hidden states or the preference of LLM itself, indicating that identifying the worst-prompt is a challenging problem. In this study, however, we get inspiration from \citet{zeng2024beear} and explore the relationship between the distance of model hidden states and the difference in model performance.

\paragraph{Setup} Specifically, we use Llama-2-13b-chat~\citep{touvron2023llama2} to generate responses to the $100$ cases in RobustAlpaca~\citep{cao2024on} with each case containing an original user query $\boldsymbol{x_0}$ and $10$ paraphrases $\boldsymbol{x_1},\boldsymbol{x_2},\ldots, \boldsymbol{x_{10}} $. To measure the quality of responses, we use ArmoRM~\citep{wang2024interpretable,wang2024arithmetic}, a top-ranked reward model in RewardBench~\citep{lambert2024rewardbench}, to give a comprehensive reward considering $19$ alignment dimensions including helpfulness, verbosity, and safety. The original query and the paraphrases are input into the Llama-2-13b-chat respectively and we obtain the hidden states after the $20$th layer (out of 40 hidden layers in total) and use the hidden state for the last token as the embedding for the sequence. Euclidean distance ($L_2$ distance) is used to measure the distance between the embeddings.  In Figure~\ref{fig:norm_distribution}, we plot the distribution of the average embedding distance between any two paraphrases of a single query (average distance) and the distribution of the distance between the original query and the worst-case paraphrasing (worst distance). In Figure~\ref{fig:norm_correlation}, we present how the worst distance correlates with the performance difference between the original query and the worst-case paraphrasing. 

\paragraph{Observation} Figure 3 shows that the worst distance displays a significantly larger mean value over the average distance (student test, $p<0.05$). Additionally, Figure~\ref{fig:norm_correlation} further verifies that the worst distance is correlated with the performance deterioration caused by the worst-case paraphrasing. The spearman's correlation coefficient is $0.36$ (student test, $p<0.05$). \textbf{Overall, the distance between a paraphrase and the original query can be an indicator for the worst-case prompt, and the worst-case prompt tends to be located in a relatively distant region from the original query.}

\subsection{Our Method: Latent Adversarial Paraphrasing}
\label{sec:lap}
\paragraph{Motivation} Intuitively, if the worst-case paraphrasing of the user query is found, we could optimize the language model on it to improve the lower bound of instruction-following performance. However, even with the general pattern observed in Section~\ref{sec:observation}, it is still a non-trivial task to construct the worst-case paraphrasing by prompting a powerful LLM since we can hardly control the embedding distance of the paraphrasing. Therefore, instead of synthesizing a human-readable textual paraphrase, we alternatively pursue a \textit{continuous paraphrase} in the latent space with the LAT framework. Namely, the perturbed hidden states could act as the latent-space paraphrasing. However, the LAT framework is unable to constrain the perturbation to preserve the semantics of the original user query. In light of this, we propose our \modelname framework. 

Specifically, in the inner loop we target enlarging the embedding distance between the original query and the \textit{continuous paraphrasing} in the latent space by maximizing the $L_2$-norm of the perturbation. Meanwhile, to preserve the original semantics, we additionally add a constraint on the increment of language modeling loss caused by the perturbation. The intuition behind this is that semantic-preserving paraphrasing generally will not increase the model perplexity or language modeling loss by a large margin~\citep{liu2024monotonic}. Mathematically, the optimization of our objective is:
\begin{equation}
\begin{aligned}
\label{eq:perturbed_loss}
    &\max_{\delta(\boldsymbol{x})} \mathbb{E}_{(\boldsymbol{x},\boldsymbol{y})  \sim \mathcal{D}} \Vert \delta(\boldsymbol{x}) \Vert_p \\
    & \rm{s.t.}\vert \mathcal{J}_{\delta(\boldsymbol{x})} - \mathcal{J}_0 \vert \leq \epsilon, \\
\end{aligned}
\end{equation}
where $\mathcal{J}_0$ and $\mathcal{J}_{\delta(x)}$ are the original loss and the loss with perturbation respectively:  
\begin{equation}
\begin{aligned}
\label{eq:original_loss}
& \mathcal{J}_0 = \mathcal{L} \left(f^\theta_{1\rightarrow L} \circ g^\theta (\boldsymbol{x}), \boldsymbol{y} \right) \\
&\mathcal{J}_{\delta(\boldsymbol{x})} = \mathcal{L}\left(f^\theta_{l+1 \rightarrow L} \left(f^\theta_{1\rightarrow l} \circ g^\theta (\boldsymbol{x}) + \delta(\boldsymbol{x}) \right), \boldsymbol{y} \right) \\
\end{aligned}
\end{equation}

During the training of the perturbation, the parameters of the language model are fixed. Then after the inner loop is optimized, the input-specific perturbation is fixed and the language model is optimized to minimize the language modeling loss with the inserted perturbation:

\begin{equation}
    \min_{\theta} \mathbb{E}_{(\boldsymbol{x},\boldsymbol{y})\sim\mathcal{D}} \mathcal{J}_{\delta(\boldsymbol{x})}
\end{equation}

\begin{figure}
    \centering
    \includegraphics[width=0.95\linewidth]{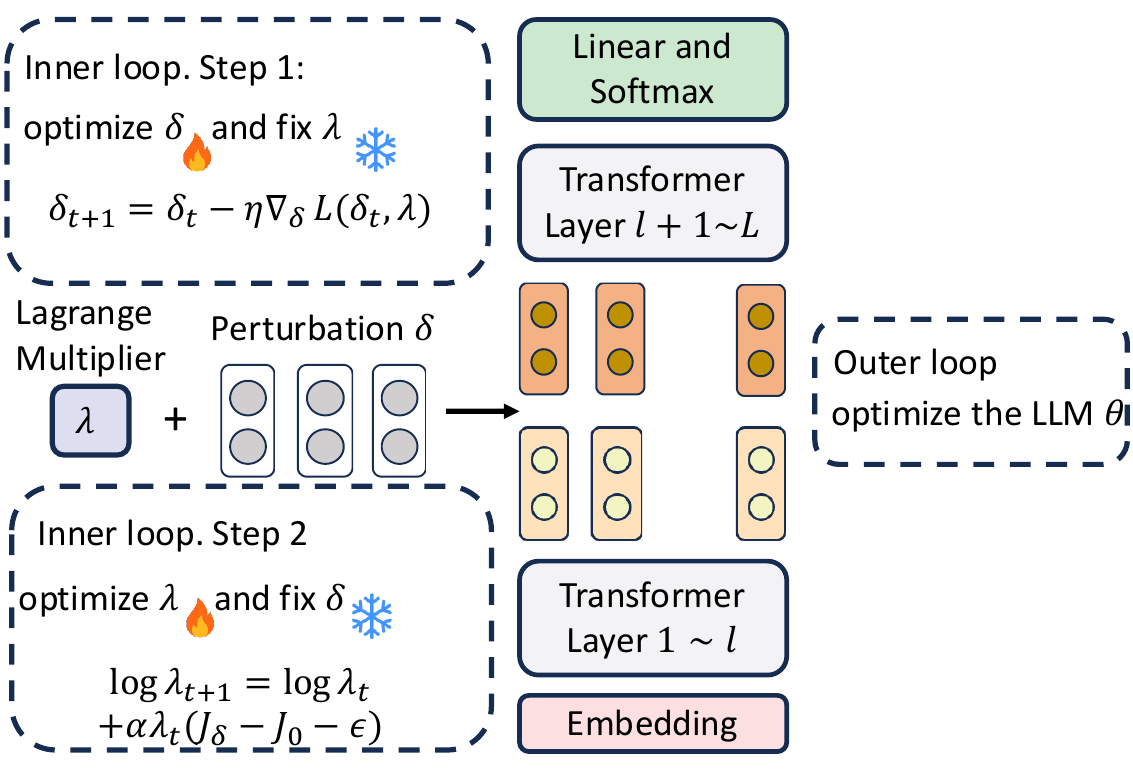}
    \caption{The workflow of our proposed \modelname framework consists of an inner loop and an outer loop. For the inner loop, two steps are conducted iteratively to update the perturbation $\delta$ and the Lagrangian multiplier $\lambda$ respectively, while for the outer loop the language model parameter $\theta$ is optimized.}
    \label{fig:workflow}
\end{figure}

\begin{algorithm}[h]
\begin{algorithmic}[1]
\STATE {\bfseries Input:} An instruction-following dataset $\mathcal{D}$, a language model $\theta$, the number of iterations in the inner loop $T$, the sampling coefficient $p$.
\FOR {$\{(\boldsymbol{x},\boldsymbol{y})\} \sim \mathcal{D}$}
    \STATE Sample $I \sim \rm{Bernouli}(p)$ 
    \IF{$I = 1$}
        \STATE **Perform Adversarial Training**
        \STATE Compute the language modeling loss with Eq~\ref{eq:original_loss}.   
        \STATE **Inner loop**
        \FOR{$j \gets 1 \mbox{~to~} T$}    
            \STATE Fix the multiplier $\lambda$ and optimize a perturbation $\delta$ using Equation~\ref{eq:optimize_delta}.
            \STATE Fix the perturbation $\delta$ and update the multiplier $\lambda$ according to Equation~\ref{eq:optimize_lambda}.
        \ENDFOR
        \STATE **Outer loop**
        \STATE Fix the perturbation $\delta$ and optimize the language model $\theta$ following Eq.~\ref{eq:perturbed_loss}. 
    \ENDIF
    \IF{ $I = 0$ \OR $p = 0$ }
        \STATE **Perform Supervised Fine-tuning**
        \STATE Optimize the language model $\theta$ on $(\boldsymbol{x},\boldsymbol{y})$ with language modeling loss following~\ref{eq:original_loss}.
    \ENDIF
\ENDFOR
\STATE {\bfseries Return: } language model $\theta$.
\end{algorithmic}
\caption{The proposed \modelname framework.}
\label{alg:alg}
\end{algorithm}

While the outer loop is straightforward, the inner loop is more complex. Different from the LAT framework, the constraint within the inner loop cannot be implemented with the projected gradient descent. Therefore, to satisfy the constraint, we leverage the Lagrangian method (an optimization technique for finding the local minima of a function over a constraint set) to convert the constrained primal inner loop problem into its unconstrained Lagrangian counterpart:
\begin{equation}
\begin{aligned}
\label{eq:lagrange_objective}
    & \mathbb{E}_{(\boldsymbol{x},\boldsymbol{y})\sim\mathcal{D}} \min_{\delta(\boldsymbol{x})} \max_{\lambda(\boldsymbol{x}) \geq 0} \mathcal{L}(\delta(\boldsymbol{x}),\lambda(\boldsymbol{x}))\\
    &\mathcal{L}(\delta(\boldsymbol{x}),\lambda(\boldsymbol{x})) = -\Vert \delta(\boldsymbol{x}) \Vert_p +\lambda (\vert \mathcal{J}_{\delta(\boldsymbol{x})} - \mathcal{J}_0 \vert - \epsilon  ), 
\end{aligned}
\end{equation}
where $\lambda \geq 0 $ serves as the Lagrange multiplier.

The constraint is transformed as a penalty term in the objective which can be dynamically modulated via the parameter $\lambda$ and the perturbation $\delta$. Specifically, to optimize the objective in Equation~\ref{eq:lagrange_objective} during the min-max game, the perturbation is updated with 
\begin{equation}
\begin{aligned}
\label{eq:optimize_delta}
& \delta(\boldsymbol{x})_{t+1} = \delta(\boldsymbol{x})_{t} - \eta \nabla_{\delta(\boldsymbol{x})} \mathcal{L} \left( \mathcal{\delta(\boldsymbol{x}), \lambda(\boldsymbol{x})} \right)\\
\end{aligned}
\end{equation}
Then the multiplier is optimized with 
\begin{equation}
\begin{aligned}
\label{eq:optimize_lambda}
& \log \lambda(\boldsymbol{x})_{t+1} = \log \lambda(\boldsymbol{x})_{t} + \alpha \lambda_{t}(\boldsymbol{x}) \left(\mathcal{J}_{\delta(\boldsymbol{x})} - \mathcal{J}_0 - \epsilon\right)
\end{aligned}
\end{equation}

Through the min-max game between the perturbation and the Lagrangian multiplier,  the transformed objective ensures that the violation of the constraint would incur a high value of $\lambda$. Since the perturbation $\delta$ and the $\lambda$ have opposite objectives, consequently, the trained perturbation is adjusted to meet the constraint.

Apart from the inner loop optimization of perturbation $\delta$ and the outer loop optimization of the language model $\theta$, we additionally introduce SFT on instruction-following data as in Equation~\ref{eq:original_loss} into our framework. We summarize the workflow of our approach in Algorithm~\ref{alg:alg}.

\section{Experiment}

\paragraph{Backbone and Dataset} We primarily adopt Llama-3-8b~\citep{dubey2024llama3} and Mistral-7b-v0.3~\citep{jiang2023mistral} as backbones for experiments. But in principle, \modelname is agnostic to the base model backbone and can be applied to any pre-trained auto-regressive language model. The models are fine-tuned on the ``chosen'' completion of \texttt{ultrafeedback\_binarized}\footnote{\scriptsize\url{https://huggingface.co/datasets/HuggingFaceH4/ultrafeedback_binarized}}, a pair-wise preference dataset preprocessed from UltraFeedback~\citep{cui2024ultrafeedback} in which the response with the highest overall score are picked as the ``chosen'' completion.

\paragraph{Benchmark and Metrics} To evaluate the prompt robustness of LLM, we leverage the RobustAlpaca benchmark~\citep{cao2024on}. Established based on the TinyAlapca~\citep{polo2024tinybenchmarks}, the benchmark contains $100$ cases with each case containing the original user query $x_0$ and $10$ semantic-preserving paraphrases $x_1,x_2,\ldots,x_{10}$ synthesized by GPT-4 and verified by human annotators. Response of GPT-4 to the original query $y_b$ serves as the baseline for computing win-rate. Specifically, $y_0,y_1,\ldots,y_{10}$ are generated based on $x_0, x_1, \ldots, x_{10}$ respectively for each cases and we use the original win-rate $p_w(x_0, y_0,y_b)$, the best win-rate $\max_i p_w (x_0,y_i,y_b)$, the worst win-rate $\min_i (x_0,y_i,y_b)$, together with the average win-rate $\frac{1}{n}\sum_i p_w(x_0,y_i,y_b)$ as our evaluation metric, where $p_w(x,y_1,y_2)$ is the probability of $y_1$ is preferred over $y_2$ given the query $x$. We use GPT-4o-mini as our evaluator for its advantage over GPT-4 in LMSYS benchmarks~\citep{chiang2023vicuna} and lower price. 

\begin{table*}[]
    \centering
    \resizebox{0.8\linewidth}{!}{
    
\begin{tabular}{lcccccccc}
\toprule
            & \multicolumn{4}{c}{Llama-3-8b}                         & \multicolumn{4}{c}{Mistral-7b-v0.3}                    \\
\cmidrule(lr){2-5}\cmidrule(lr){6-9}
            & \makecell[c]{original\\ win-rate} & \makecell[c]{best \\win-rate} & \makecell[c]{worst\\ win-rate} & \makecell[c]{average \\win-rate} & \makecell[c]{original \\ win-rate} & \makecell[c]{best \\win-rate} & \makecell[c]{worst\\ win-rate} & \makecell[c]{average \\win-rate} \\
\midrule
\multicolumn{9}{l}{\textit{Training algorithms}} \\
SFT         & \textbf{15.06}        & \underline{47.24}       & 1.44        & \underline{15.39}       & 7.85         & \underline{43.62}       & 0.01        & \underline{12.62}       \\
Data Augmentation          & 11.47        & 39.78       & 1.77        & 13.75       & 9.67         & \textbf{44.33}       & \underline{1.64}        & 12.38       \\
Prompt Consistency          & 10.78        & 37.76       & \textbf{3.09}        & 11.64        & 8.68         & 42.85       & 1.00        & 10.50       \\
LAT	& 10.97	& 45.23	& \underline{2.04} & 14.98 & \underline{11.02} & 37.32 & 1.38 & 12.06 \\
\hdashline
\multicolumn{9}{l}{\textit{Inference algorithms}} \\
USC         & 8.14         & 45.91       & 1.01        & 15.22       & 6.47         & 39.3        & \textbf{1.85}        & 12.25       \\
Self-Refinement & 4.02         & 28.82       & 1.00        & 6.74        & 9.96        & 29.82       & 0.00           & 7.40        \\
Mixture-of-Agents         & 6.19         & 40.33       & 0.00           & 12.19       & 5.51         & 32.04       & 0.00           & 8.18   \\
\midrule
\modelname  &	\underline{12.56} & \textbf{49.79} & 1.99 & \textbf{15.48} & \textbf{11.05} & 41.93 &1.59 & \textbf{14.05} \\
\bottomrule
\end{tabular}

    }
    \caption{Evaluation performance on RobustAlpaca~\citep{cao2024on} with Llama-3-8b and Mistrail-7b-v0.3. The numbers in bold are the best results and the numbers underlined are the second best results.} 
    \label{tab:main}
\end{table*}

\paragraph{Baseline Method} To verify the effectiveness of the proposed \modelname, we draw a comparison with existing baseline methods. Apart from \textbf{vanilla SFT}, we also consider:
\begin{itemize}
    \item \textbf{Data Augmentation} diversities the instruction-following data with augmented paraphrases.  Specifically, we sample 10k queries from UltraFeedback, synthesize a paraphrase for each query with a proprietary LLM, and augment the synthesized paraphrases into the instruction-following data. The template for prompting paraphrase generation is borrowed from \citet{cao2024on}. 
    \item \textbf{Prompt Consistency}~\citep{zhou2022prompt} design a loss function to minimize the divergence of model outputs on different paraphrases.  In our implementation, we reuse the $10$k augmented paraphrases and distill the model response from one query to its paraphrase.  
    \item \textbf{Latent Adversarial Training (LAT)}~\citep{casper2024defending} includes a trainable perturbation into the hidden states and optimize the language model parameters to minimize the language modeling loss at the existence of perturbation.  
\end{itemize}
Meanwhile, we also compare our approach with several inference-time algorithms including 
\begin{itemize}
    \item \textbf{Self-Refinement}~\citep{cao2024on} prompts the SFT-ed language model to first rewrite the query according to its own preference before generating a response to the query.   
    \item \textbf{Universal Self-Consistency (USC)}~\citep{chen2024universal} first generates multiple draft responses with different hyper-parameter settings using the SFT-ed language model and then prompts the same SFT-ed language model itself to choose the final output from the candidates. We generate $4$ draft responses using different random seeds in out implementation.    
    \item \textbf{Mixture-of-Agents}~\citep{anonymous2025mixtureofagents} is a recently proposed inference algorithm with layers of multiple agents and each agent provides a response given all the draft responses generated by the agents in the previous layers. We construct a 2-layer mixture-of-agents network with each layer comprising $4$ SFT-ed models.
\end{itemize}

\paragraph{Experimental Results} We report the evaluation performance of our proposed \modelname and the baseline methods on Table~\ref{tab:main}. From the table, we can observe that 
(1) Our approach obtains the best results or the second best result for the original win-rate and the average win-rate on two backbones, verifying that our approach enhances model robustness to different paraphrasing of user queries.   
(2) Contrary to recent findings on the importance of inference-time scaling, the inference-time algorithms in the baseline methods do not exhibit an advantage over vanilla SFT. Notably, Mixture-of-Agents and Self-Refinement have the lowest results on the worst win-rate, possibly because they both rely on external capable LLM to aggregate the strengths and weaknesses of draft responses. 
(3) Even with the baseline methods or our proposed approach, there is still a large margin between the best win-rate and the worst win-rate, suggesting a large room for further improvement.

\section{Analysis}
\begin{table}[]
    \centering
    \resizebox{0.8\linewidth}{!}{
    
\begin{tabular}{lcccc}
\toprule
             & \makecell[c]{original \\ win-rate} & \makecell[c]{best \\win-rate} & \makecell[c]{worst \\win-rate} & \makecell[c]{average \\win-rate} \\
\midrule
\modelname      & 10.21        & \textbf{53.39}       & \textbf{4.98}        & \textbf{17.94}       \\
\midrule
\quad $p=0$ & 14.52        & 48.55       & 4.67        & 17.49       \\
\quad $p=1$  & 5.04         & 46.44       & 0.30        & 11.42       \\
\quad $\lambda = 0$         & \textbf{10.42}        & 45.81       & 1.56        & 16.68      \\
\midrule
\end{tabular}
    }
    \caption{Ablation experiments on RobustAlpaca with Llama-3-8b backbone. Numbers in bold are best results.}
    \label{tab:ablation}
\end{table}
\paragraph{Ablation Study} To evaluate the importance of each module in our \modelname framework and how they contribute to the overall performance, we perform an ablation study on the following variants: (1)$\lambda = 0$: we fix the value of $\lambda$ to be zero and thus invalidate the constraint on language modeling loss; (2) $p=1$: the supervised fine-tuning (line 15 in Algorithm~\ref{alg:alg}) is removed from our framework; (3) $p=0$: we do multi-tasking (line 15 in Algorithm~\ref{alg:alg}) and use every datapoint for both adversarial training and SFT.

\begin{table*}[h]
    \centering
    \resizebox{0.75\linewidth}{!}{
        
\begin{tabular}{lccccccc}
\toprule
     & MMLU  & ARC-E & ARC-C & Hellaswag & Winogrande & TruthfulQA & Average \\
\midrule
SFT  & 61.86 & 80.30    & 50.17         & 60.13     & 74.27      & 32.44      & 59.86   \\
Data Augmentation   & 62.08 & 80.64    & 50.09         & 60.07     & 74.66      & 31.82      & 59.89   \\
Prompt Consistency   & 62.28 & 80.81    & 50.60         & 60.10     & 74.66      & 32.44      & 60.15   \\
LAT & 61.61 & 62.84 & 49.57 & 60.00 & 74.51 & 33.17 & 56.95 \\
\midrule
\modelname & 61.92 & 81.02    & 52.13         & 60.38     & 74.66      & 32.44      & 60.43   \\
\midrule
SFT  & 58.79 & 78.32    & 45.39         & 60.96     & 74.98      & 33.90      & 58.72   \\
Data Augmentation    & 58.81 & 78.49    & 45.65         & 60.74     & 74.74      & 33.54      & 58.66   \\
Prompt Consistency   & 59.24 & 78.24    & 43.94         & 60.75     & 74.51      & 33.05      & 58.29   \\
LAT & 57.42 & 76.94 & 45.14 & 59.98 & 74.35 & 30.97 & 57.47 \\
\midrule
\modelname & 58.72 & 77.36    & 44.97         & 60.74     & 73.95      & 34.39      & 58.36 \\
\bottomrule
\end{tabular}
    }
    \caption{Down-stream task evaluation results on Llama-3-8b (upper block) and Mistral-7b-v0.3 (bottom block). Our \modelname preserves the performance on basic commonsense and knowledge as the results are similar.} 
    \label{tab:benchmark}
\end{table*}

The experiment results on Llama-3-8b is shown in Table~\ref{tab:ablation}. From the table, we can observe that our \modelname obviously outperforms the $\lambda = 0$ variants in most metrics, suggesting the necessity of the constraint on language modeling loss. Moreover, compared with the $p=0$ variant that all data points are trained with  SFT and the $p=1$ variant that eliminates the SFT process, \modelname obtains a better balance between objectives.

\begin{table}[]
    \centering
    \resizebox{0.9\linewidth}{!}{
        \begin{tabular}{lccc}
\toprule
                  & 1st turn & 2nd turn & Average \\
\midrule
SFT               & 5.34     & 4.04     & 4.69    \\
Data Augmentation & 5.26     & 3.96     & 4.61    \\
Prompt Consistency & 5.15     & 4.10     & 4.63    \\
LAT               & 5.19     & 4.25     & 4.72    \\
\midrule
\modelname            & 5.11     & 4.23     & 4.67    \\
\bottomrule
\end{tabular}
    }
    \caption{Evaluation results on MT-bench~\citep{chiang2023vicuna} using Llama-3-8b backbone. Our \modelname preserves the performance on instruction-following as the results are similar.}
    \label{tab:mt_bench}
\end{table}

\paragraph{Downstream Task Performance} To examine how our robustness enhancement approach affects downstream task performance, we evaluate the capacity of \modelname on several benchmarks from huggingface open LLM leaderboard in comparison with baseline training algorithms. The evaluation results are shown in Table~\ref{tab:benchmark}. Additionally, we examine and compare their instruction-following ability on MT-bench~\citep{chiang2023vicuna}, with the evaluation results presented in Table~\ref{tab:mt_bench}. We use GPT-4o-mini as evaluator on MT-bench. From the two tables, we can observe that most baseline training algorithms will not result in a drastic change either downstream tasks or instruction-following benchmarks. As \modelname is on par with and sometimes outperforms the baseline training algorithms, it is therefore verified that the improvement in robustness brought by our approach is not at the cost of performance deterioration on basic commonsense or instruction-following ability.

\begin{table}[]
    \centering
    \resizebox{0.95\linewidth}{!}{
        \begin{tabular}{lcccc}
\toprule
         & \makecell[c]{original \\win-rate} & \makecell[c]{best\\ win-rate} & \makecell[c]{worst\\ win-rate} & \makecell[c]{average\\ win-rate} \\
\midrule
\multicolumn{5}{l}{\quad \textit{Training algorithms}} \\
SFT      & \underline{12.47}        & 37.03       & 0.04        & 13.24       \\
Data Augmentation & 8.43         & \underline{40.36}       & \underline{2.50} & 13.34       \\
Prompt Consistency & 14.80        & 39.73       & 1.36        & 12.99       \\
LAT      & 10.01        & 35.61       & \underline{2.50}        & 13.81       \\
\hdashline
\multicolumn{5}{l}{\quad \textit{Inference algorithms}} \\
USC      & 6.74         & \textbf{44.95}       & 1.77        & \textbf{14.81}       \\
Self-Refinement   & 2.55         & 23.49       & \underline{2.50}        & 5.64        \\
Mixture-of-Agents      & 5.47         & 36.96       & 0.00        & 7.82        \\
\midrule
\modelname   & \textbf{13.31}        & 37.08       & \textbf{4.01}        & \underline{14.08} \\
\bottomrule
\end{tabular}
    }
    \caption{Evaluation performance on RobustAlpaca~\citep{cao2024on} with Llama-2-13b. The numbers in bold are the best results and the numbers underlined are the second best results.}
    \label{tab:llama-2-13b}
\end{table}

\paragraph{Scalability on Larger Backbones} To explore whether the \modelname framework could work on larger LLM backbones, we perform the experiments on Llama-2-13b~\citep{touvron2023llama2} and evaluate their performance using $40$ cases randomly sampled from RobustAlpaca. The experimental results are presented in Table~\ref{tab:llama-2-13b}. As we can observe from the table, \modelname obtains the best result and the second-best results on the original win-rate and the average win-rate respectively, suggesting that our approach remains effective on larger backbones. Meanwhile, it is worth noting that USC is a competitive baseline for Llama-2-13b,  different from the experiment results on smaller models in Table~\ref{tab:main}. We gauge that Llama-2-13b is more capable in summarizing and aggregating candidate answers and therefore is able to obtain better performance with USC.

\begin{figure*}
\centering
\begin{subfigure}{0.33\textwidth}
\includegraphics[width=\textwidth]{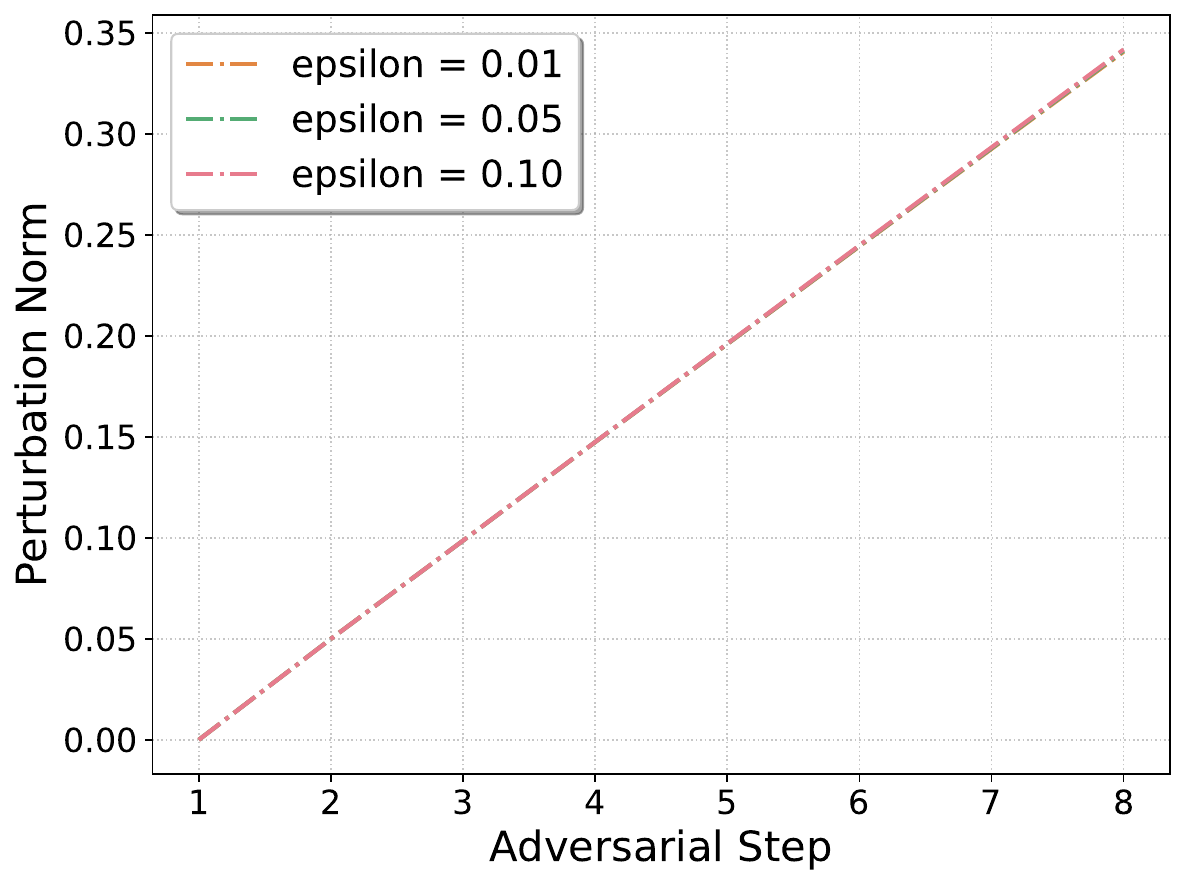}
\caption{Dynamics of perturbation norm $\Vert\delta\Vert_2$ }
\label{fig:training_delta}
\end{subfigure}
\begin{subfigure}{0.33\textwidth}
\includegraphics[width=\textwidth]{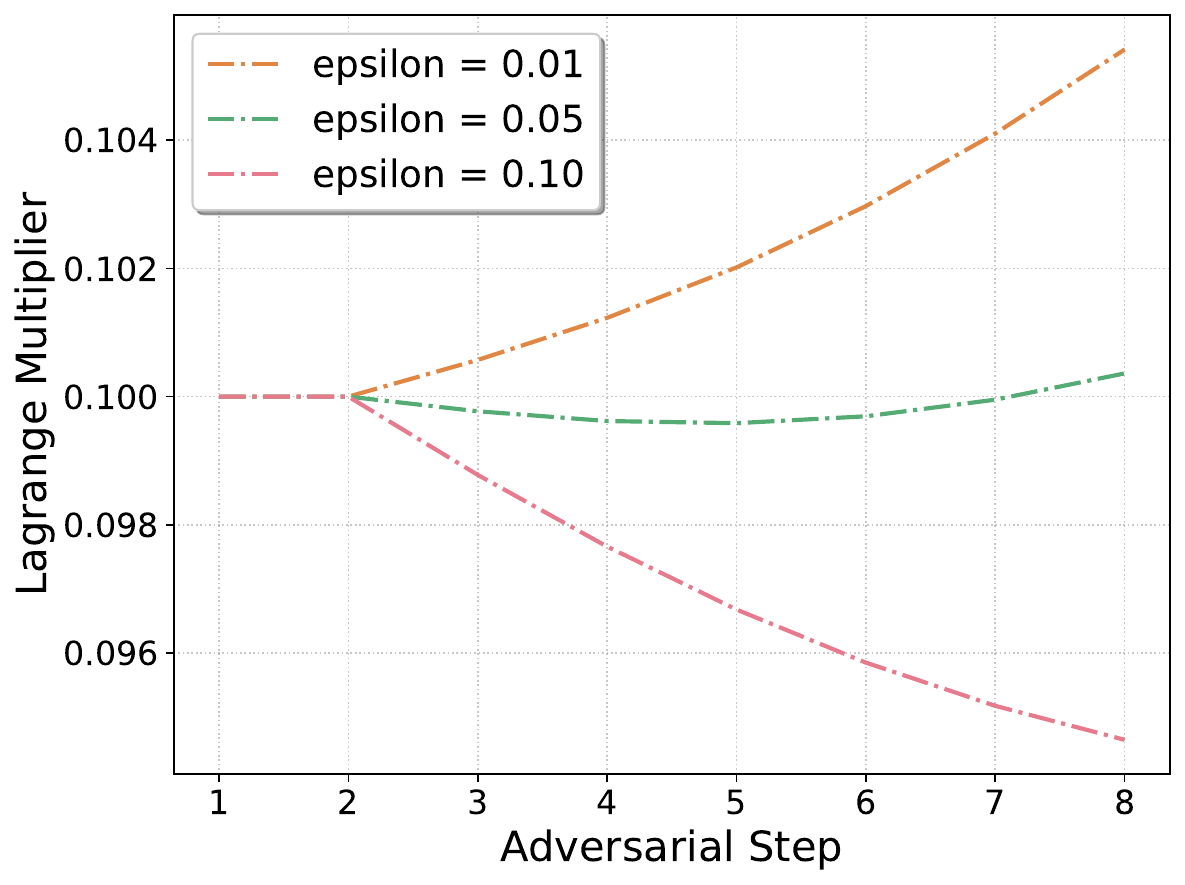}
\caption{ Dynamics of Lagrange multiplier  $\lambda$ }
\label{fig:training_lambda}
\end{subfigure}
\begin{subfigure}{0.33\textwidth}
\includegraphics[width=\textwidth]{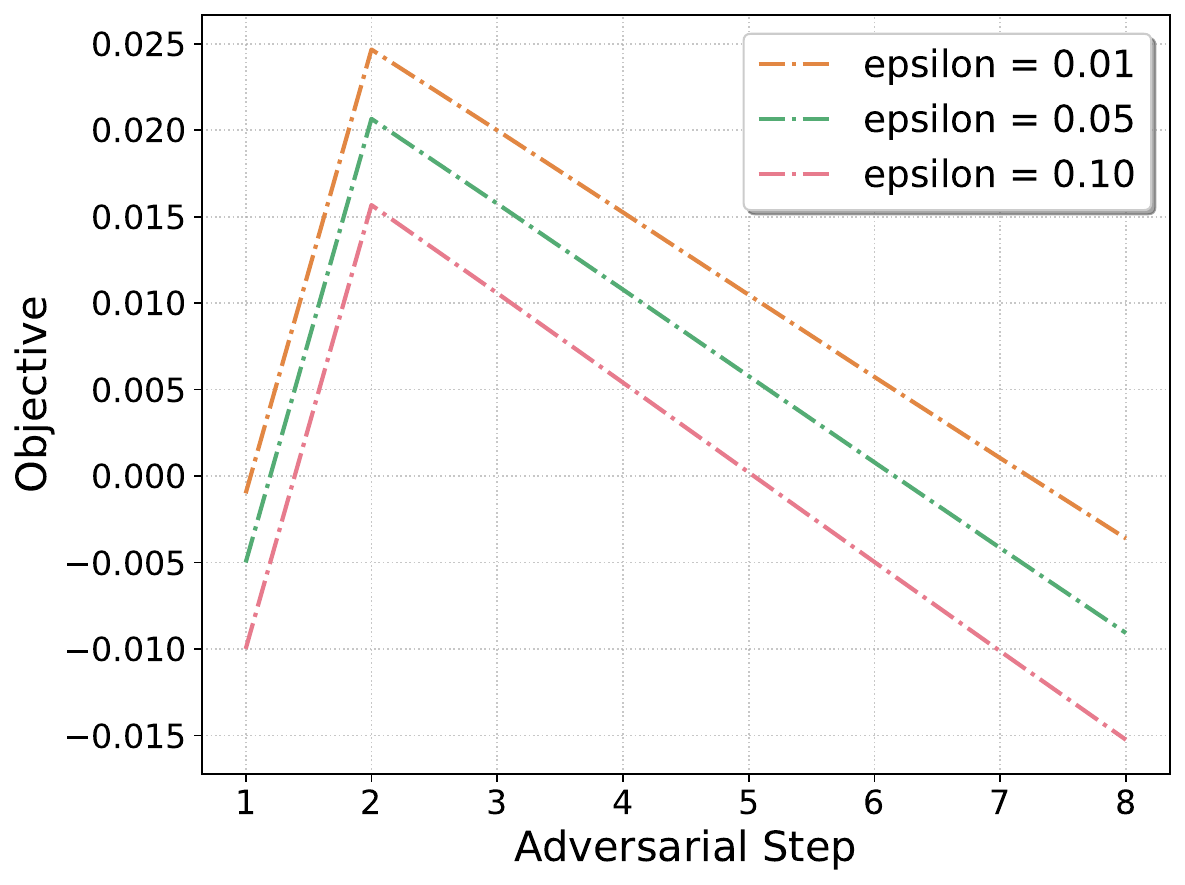}
\caption{Dynamics of $\mathcal{L}(\delta(\boldsymbol{x}),\lambda(\boldsymbol{x}))$ }
\label{fig:training_objective}
\end{subfigure}
\caption{Training dynamics for the inner-loop optimization with different constraint margin $\epsilon$ on Llama-3-8b backbone. }
\label{fig:inner_dynamics}
\end{figure*}

\begin{figure}
    \centering
    \includegraphics[width=0.90\linewidth]{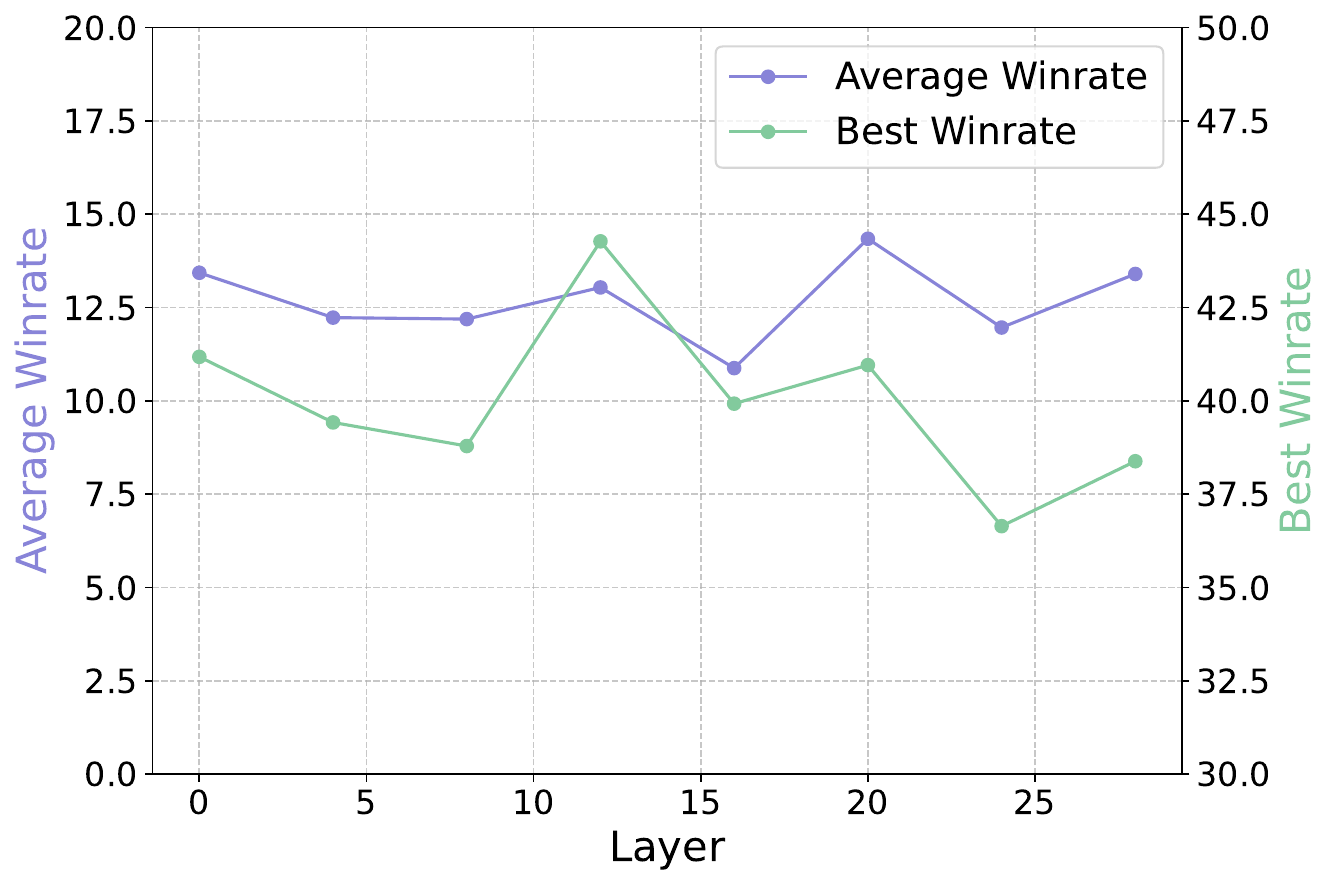}
    \caption{The trend of average win-rate and the best win-rate with the perturbation position on Mistral-7b-v0.3 backbone.}
    \label{fig:layer_trend}
\end{figure}

\paragraph{Effect of Perturbation Position} To understand how the position of the perturbation (i.e., the transformer layer at which we incorporate the perturbation) impacts the effectiveness of our approach, we vary the positions of the perturbation among transformer layers in Mistral-7b-v0.3~\citep{jiang2023mistral} and plot the trend of the best win-rate and the average win-rate in Figure~\ref{fig:layer_trend}. From the figure, it seems that Layer 12 and Layer 20 are slightly better than other layers as suitable positions for perturbation as the \modelname optimization at the two positions yields the maximum best win-rate and the maximum average win-rate respectively. But overall, the performance on average win-rate is relatively insensitive to the position of the perturbation. In this study the perturbation is only inserted into one layer following previous works in LAT~\cite{casper2024defending,sheshadri2024latent}. We leave the multi-layer perturbation to future works.

\paragraph{Analysis on Training Dynamics.} To gain insight into the process of inner-loop optimization, we analyze the iteration of multiple variables within the training objective of Equation~\ref{eq:lagrange_objective} and plot their trend in Figure~\ref{fig:inner_dynamics}. Specifically, for different values of the constraint margin ($\epsilon= 0.01,\ 0.05,\ 0.10$), we record the perturbation norm $\Vert \delta\Vert_2$, Lagrange multiplier $\lambda$ and the objective $\mathcal{L(\delta(\boldsymbol{x}),\lambda(\boldsymbol{x}))}$ for each inner-loop iteration, and plot the trend of their average value among the instruction-following dataset. Notably, a smaller value of $\epsilon$ means a more tight constraint. As we can observe from Figure~\ref{fig:inner_dynamics}, the strength of the constraint has little effect on the perturbation norm as the perturbation norm steadily grows during the inner-loop training process whatever the value of $\epsilon$ is. However, the trend of Lagrange multiplier $\lambda$ is obviously influenced by the $\epsilon$ since there is a monotonic decrease of $\lambda$ when setting $\epsilon = 0.10$ but an opposite trend of $\lambda$ could be observed when setting $\epsilon = 0.01$. We gauge the reason behind is that a tight constraint is harder to satisfy and thus pushes the Lagrange multiplier to be larger. It could also explain why a larger value of $\epsilon$ corresponds to a larger inner-loop objective in Figure~\ref{fig:training_objective} since the difference in objectives is dominated by the second term or the penalty term.

\begin{table}[]
    \centering
    \resizebox{0.85\linewidth}{!}{
        \begin{tabular}{lcccc}
\toprule
           & \makecell[c]{original \\ winrate} & \makecell[c]{best \\ winrate} & \makecell[c]{worst \\winrate} & \makecell[c]{average \\ winrate} \\

\midrule
SFT+DPO    & 22.95        & 60.95       & 4.83        & 24.61       \\
\modelname+DPO & 23.16        & 71.20       & 5.12        & 31.53       \\
\bottomrule
\end{tabular}
    }
    \caption{The evaluation performance on RobustAlpaca with subsequent preference learning on Llama-3-8b backbone.}
    \label{tab:dpo}
\end{table}

\paragraph{Compatibility with Preference Learning.} To examine whether the proposed \modelname framework is compatible with existing preference learning techniques, we perform direct preference optimization (DPO)~\citep{rafailov2023direct} on SFT-ed model and the \modelname-ed model to draw a comparison. As we can observe from the table, with subsequent preference learning our approach outperforms vanilla SFT by a large margin, verifying that our approach does not interfere with and is compatible with the subsequence preference learning.

\section{Conclusion}
In this study, we tackled the challenge of prompt robustness in LLMs, particularly their sensitivity to paraphrased user queries. Our findings revealed that the Euclidean distance between a paraphrased query and the original input correlates with response quality. Leveraging this insight, we introduced \modelname, a novel adversarial framework that continuously searches for latent paraphrases and optimizes LLM parameters to enhance robustness. Through extensive experiments across multiple model architectures, we demonstrated that our approach consistently improves performance across paraphrases without compromising commonsense reasoning. Future work will explore extending \modelname~to more models and datasets, investigating its applicability to multi-turn dialogues, and refining adversarial training techniques to further enhance model resilience. 

\section*{Acknowledgments}
We thank Stephen Casper, Javier Rando, Ryuto Koike and Samuele Marro for feedback and comments on an earlier draft. 

\section*{Impact Statements}
This work on improving prompt robustness through \modelname has significant implications for the reliable deployment of large language models in real-world applications. By addressing the fundamental challenge of prompt sensitivity, this approach could enhance the consistency and reliability of LLM outputs across different but semantically equivalent user queries. This improvement is particularly crucial for high-stakes applications such as healthcare, legal assistance, and educational systems, where inconsistent responses could lead to harmful outcomes. However, the development of more robust LLMs also raises important ethical considerations. While increased robustness could help prevent accidental failures and reduce unwanted biases triggered by prompt variations, it might also make models more consistently capable of harmful outputs if misused.

\clearpage

\bibliography{example_paper}
\bibliographystyle{icml2025}

\newpage
\appendix
\onecolumn

\section{Hyper-parameter Setting}
\label{app:impl}

\begin{table}[]
    \centering
    \resizebox{0.5\linewidth}{!}{
    \begin{tabular}{lccc}
\toprule
                        & Llama-3-8b                    & Mistral-7b-v0.3              & Llama-2-13b                              \\
\midrule
Precision               & \texttt{bfloat16}                         & \texttt{bfloat16}                         & \texttt{bfloat16}                         \\
max sequence length     & 1024                              & 1024                              & 1024                              \\
Batch size              & 32                               & 32                               & 32                               \\
Optimizer               & AdamW                            & AdamW                            & AdamW                            \\
Adam ($\beta_1$, $\beta_2$) & $(0.9, 0.95)$                      & $(0.9, 0.95)$                      & $(0.9, 0.95)$                     \\
Learning rate           & 1e-4                         & 1e-4                         & 1e-4                         \\
Warmup ratio            & 0.1                              & 0.1                              & 0.1                              \\
Decay style             & \texttt{cosine}                           & \texttt{cosine}                           & \texttt{cosine}                           \\
Weight decay            & 0.0                                & 0.0                                & 0.0                                \\
Training step           & 1 epoch                          &  1 epoch                        & 1 epoch                          \\
LoRA rank               & 64                               & 64                               & 64                               \\
LoRA alpha              & 16                               & 16                               & 16                               \\
LoRA dropout            & 0.05                             & 0.05                             & 0.05                             \\
LoRA modules          &  \texttt{\makecell[c]{gate\_proj,\\up\_proj,\\down\_proj}}  &  \texttt{\makecell[c]{gate\_proj,\\up\_proj,\\down\_proj}} &  \texttt{\makecell[c]{gate\_proj,\\up\_proj,\\down\_proj}}\\
\bottomrule
\end{tabular}
    }
    \caption{Hyper-parameter settings for supervised fine-tuning and preference learning.}
    \label{tab:hyper_parameter}
\end{table}

Our experiments are conducted on a cloud Linux server with Ubuntu 16.04 operating system. The codes are written in Python 3.10 with the huggingface libraries\footnote{\scriptsize\url{https://github.com/huggingface/trl}}. 
We run our experiments on Nvidia Tesla A100.  The detailed hyper-parameter settings for different backbones are shown in Table~\ref{tab:hyper_parameter}, which mostly follows \citet{lee2023platypus} and \citet{ivison2023camels}. At inference, we use nucleus sampling with $p=0.9$ and temperature $T=1.0$. vLLM~\footnote{\scriptsize\url{https://github.com/vllm-project/vllm}} is adopted for accelerating response generation.

\section{More Details About Baseline Implementation}
\label{app:baseline}
In this section, we elaborate on the template used for inference-based algorithms and the template for synthesizing paraphrases of user queries.  
Specifically, we adopt the template from \citet{cao2024on} to produce semantic-preserving paraphrases and the template is shown below:
\clearpage

\noindent\fbox{
    \parbox{\linewidth}{
{\itshape
Your task is to generate one unique paraphrase for the given query, ensuring that the meaning of the paraphrase remains consistent, and the structure is significantly altered. Do not introduce any new information that isn't present in the original query, and avoid omitting any crucial information from the original query, particularly any specific requirements about the output content, style, format, options, or any numbers or data. The responses to your paraphrases and the original query should be identical. If the original query contains quoted or referenced content, such as 'rewrite the following sentence: '$<$reference$>$' and those enclosed in '$<$br$><$br$>$', you should retain this referenced content in your paraphrases and rephrase the rest of the text. Your task is not to answer the query, but solely to rephrase it. Please provide unique and creative rephrasing. 

The examples provided below are for illustrative purposes only. 

Examples:
\newline
\newline
Query: ``Why is it that only proteins, carbohydrates and fats are deemed to have caloric value?$<$br$>$Just to clarify - consider the hypothetical situation below:$<$br$>$100g of 'carbohydrates' = 400kCal$<$br$>$100g of 'protein' = 400kCal$<$br$>$100g of 'fat' = 900kCal$<$br$>$100g of 'vitamins, minerals and/or other electrolytes' = ?''
\newline
\newline
Output: Why are proteins, carbohydrates, and fats the only substances considered to have caloric value? To provide further clarification, let’s consider the hypothetical scenario below: If 100g of carbohydrates provide 400kCal, 100g of protein provides 400kCal, and 100g of fat provides 900kCal, what would be the caloric value of 100g of vitamins, minerals, and/or other electrolytes?
\newline
\newline
Query: ``I have this text: $<$br$><$br$>$We use professional processes. Use the photo delivery system, check the quality and send it to the customer immediately. We started editing digital photos from the first wave of 2000. Started with editing personal photos, and then spread our passion by editing wedding photos or restoring old photos for loved ones, friends and family. friends and customers. 2005 started a local photo shop and started commercial photo editing, and product photography for manufacturers. 2007 implemented online projects on freelancer platforms.$<$br$><$br$>$Rewrite it in professional style to tell a touching story.''
\newline
\newline
Output: I possess this text: <br><br>We use professional processes. Use the photo delivery system, check the quality and send it to the customer immediately. We started editing digital photos from the first wave of 2000. Started with editing personal photos, and then spread our passion by editing wedding photos or restoring old photos for loved ones, friends and family. friends and customers. 2005 started a local photo shop and started commercial photo editing, product photography for manufacturers. 2007 implemented online projects on freelancer platforms.<br><br>Rephrase it in a professional manner to narrate a moving story.
\newline
\newline
Query: ``I want you to act as a Tiktok Ads Keywords generator. Your task is to come up with keywords that can be used in Tiktok advertising campaigns. The keywords should be relevant to the target audience and the product or service being advertised. Your responses should not include more than 5 keywords per request.''
\newline
\newline
Output: Your role will be to serve as a TikTok Ads Keywords generator. Your objective is to generate relevant keywords that can be utilized in TikTok advertising campaigns. These keywords should effectively target the desired audience and align with the product or service being promoted. Please provide no more than five keywords per request in your responses.
\newline
\newline
Now, please generate a paraphrase for the given query.
\newline
\newline
Query: \{query\}

Output:
}}}
\newline
\newline

The prompt for Self-Refinement~\citep{cao2024on} is shown below:
\newline
\newline
\noindent\fbox{
    \parbox{\linewidth}{
{\itshape
Given the following instruction: \{query\}, paraphrase the instruction to be more natural and then generate a response. Start the paraphrase with ``Paraphrase: '' and start your response with ``Response: '' (without quotes)
}}}
\newline
\newline

The prompt for USC~\citep{chen2024universal} is shown below:
\newline
\newline
\noindent\fbox{
    \parbox{\linewidth}{
{\itshape
Evaluate these responses. 

Select the most consistent response based on majority consensus.

Answer with ``The most consistent response is Response X'' (without quotes).
\newline
\newline
Response 1: \{draft response\}

Response 2: \{draft response\}

Response 3: \{draft response\}

Response 4: \{draft response\}
}}}
\newline
\newline

The prompt for Mixture-of-Agents~\citep{anonymous2025mixtureofagents} are shown below:
\newline
\newline
\noindent\fbox{
    \parbox{\linewidth}{
{\itshape
You have been provided with a set of responses from various open-source models to the latest user query. Your task is to synthesize these responses into a single, high-quality response. It is crucial to critically evaluate the information provided in these responses, recognizing that some of it may be biased or incorrect. Your response should not simply replicate the given answers but should offer a refined, accurate, and comprehensive reply to the instruction. Ensure your response is well-structured, coherent, and adheres to the highest standards of
accuracy and reliability.
\newline
\newline
Responses from models:

Response 1: \{draft response\}

Response 2: \{draft response\}

Response 3: \{draft response\}

Response 4: \{draft response\}

Output the synthesized response directly without any prefix.

}}}
\newline
\newline




\end{document}